\documentclass{article}

\usepackage{PRIMEarxiv}

\usepackage[utf8]{inputenc} 
\usepackage[T1]{fontenc}    
\usepackage{hyperref}       
\usepackage{url}            
\usepackage{booktabs}       
\usepackage{amsfonts}       
\usepackage{nicefrac}       
\usepackage{microtype}      
\usepackage{lipsum}
\usepackage{fancyhdr}       
\usepackage{graphicx}       
\graphicspath{{media/}}     
\usepackage{amsmath}
\usepackage{enumitem}

\title{Personalized Federated Learning via Stacking
}

\author{
  Emilio Cantu-Cervini \\
  ecantuc@umich.edu \\
}

\begin{document}
\maketitle

\begin{abstract}
Traditional Federated Learning (FL) methods typically train a single global model collaboratively without exchanging raw data. In contrast, Personalized Federated Learning (PFL) techniques aim to create multiple models that are better tailored to individual clients' data. We present a novel personalization approach based on stacked generalization where clients directly send each other privacy-preserving models to be used as base models to train a meta-model on private data. Our approach is flexible, accommodating various privacy-preserving techniques and model types, and can be applied in horizontal, hybrid, and vertically partitioned federations. Additionally, it offers a natural mechanism for assessing each client's contribution to the federation. Through comprehensive evaluations across diverse simulated data heterogeneity scenarios, we showcase the effectiveness of our method.
\end{abstract}


\section{Introduction}
Federated Learning (FL) is an area of research that develops methods to allow multiple parties to collaboratively train machine learning models without exchanging data. First introduced in 2016 by McMahan et al. to allow a large number of edge devices to collaboratively train language models \cite{mcmahan2023communicationefficient}, FL has been successfully applied to several domains where for regulatory or privacy reasons models cannot be trained on centralized pooled data.

Most FL approaches result in a single collaboratively trained global model that is used by every client for inference. Personalized Federated Learning (PFL) recognizes that in some non-IID contexts performance improvements are possible if each client somehow adapts or personalizes the global model to its data. Approaches range from clients fine-tuning the global model on private data to client clustering, and others discussed in Section \ref{sec:related}.

In this paper, we build on prior work \cite{wu_wrapperfl:_2022} and explore a simple personalization approach that avoids training a global model which is then personalized. Instead, clients employ privacy-preserving techniques \cite{pp-survey} to train a model on their data and make it public to the federation. Each client then fetches public models and uses them in addition to a private (non-privacy-preserving) model as base predictors to train a meta-model via stacked generalization \cite{stacked-gen} with private data.

\begin{figure}
  \centering
  \makebox[\textwidth]{\includegraphics[width=0.8\textwidth]{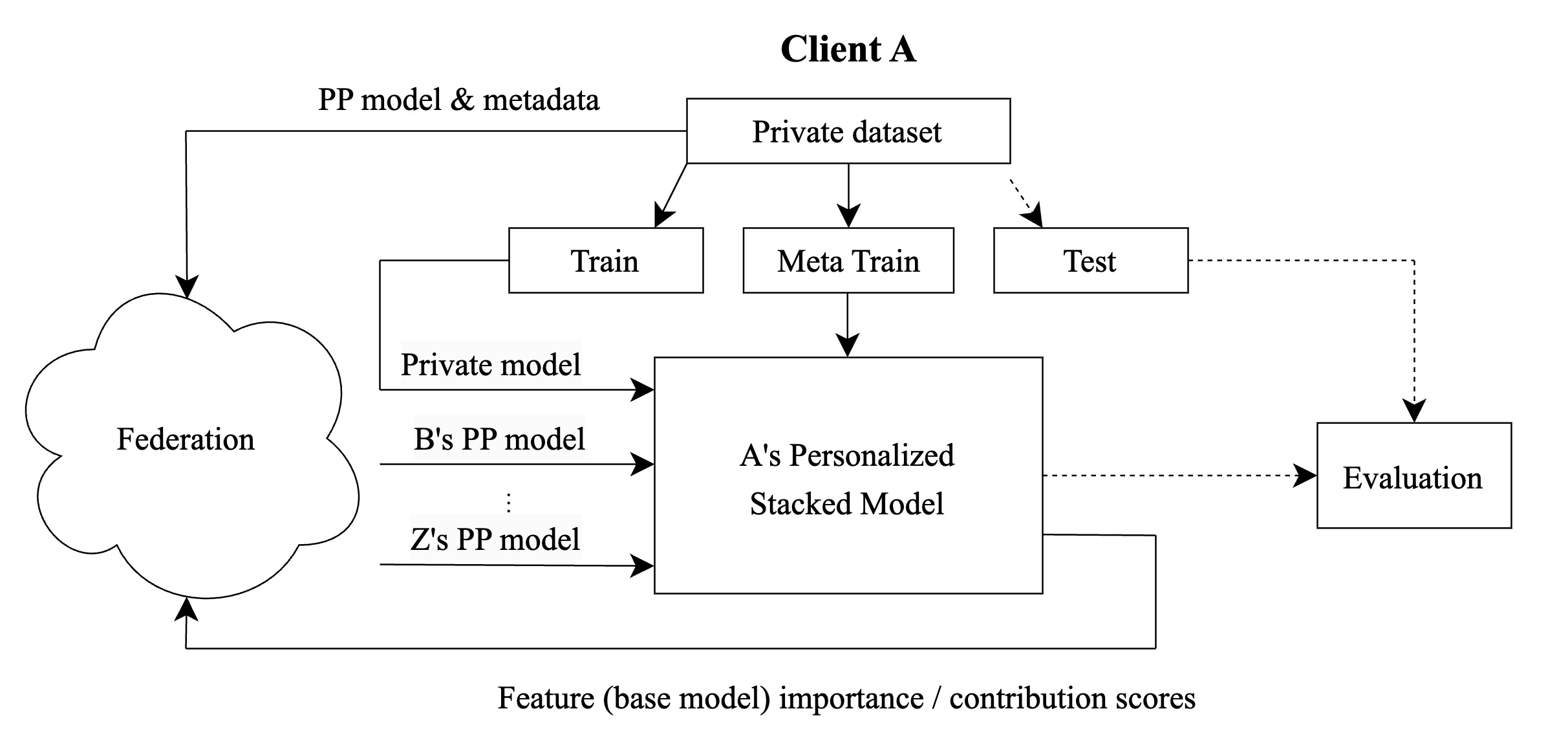}}
  \caption{A high level outline of the proposed personalization approach. A client shares a privacy-preserving (PP) model trained on private data along with metadata with the federation and trains a personalized stacked model with its own private model and those fetched from the federation. It can then use feature importance methods to rank or score the contributions of each retrieved model and report it to the federation. In our experiments each personalized models is evaluated on a held-out test set of each client's private data.}
  \label{fig:method_fig}
\end{figure}

Intuitively, the stacked model learns to weigh and combine other clients' models to best predict its own data. Thus, even if every client has access to the same base models, every stacked model will be personalized to each client by training it on private data. Our approach is summarized in Figure \ref{fig:method_fig} and discussed in more detail in Section \ref{sec:method}. Although simple, we highlight some immediate advantages of our proposal:

\begin{itemize}[leftmargin=*]
\item \textbf{Flexibility.} Since stacked generalization is model agnostic, clients can freely choose the model type and privacy-preserving technique applied to their public model as long as prediction is possible by other clients. Thus for example, while some clients could opt for sharing the unmodified weights of their linear model, others might opt for Differential Privacy \cite{diff_priv} approaches with different privacy budgets or Homomorphic Encryption \cite{homo_enc} schemes.

\item \textbf{Simplified aggregator logic.} Most previous personalization approaches require an aggregator to coordinate the global model's training (combining gradients, dealing with stragglers, etc). In our method, the aggregator's logic is greatly simplified as it need only act as a model database and perhaps maintain a "contribution graph" (explained in Section 3). 

\item \textbf{Natural contribution evaluation.} Applying standard variable importance methods to the personalized stacked models is a natural way to estimate how much each retrieved model contributed to a client’s predictive performance. The resulting importance rankings or scores can then be used to implement fairness mechanisms. For example, clients can aggregate their scores into a Contribution Evaluation \cite{siomos2023contribution} metric that attempts to capture each client's overall contribution to the federation.

\item \textbf{Data partitioning.} As we will discuss in Section \ref{sec:method}, our method is easily adapted to vertical and hybrid data partitioning scenarios where clients have potentially different sets of features. 
\end{itemize}

In addition, the method's main shortcomings include:

\begin{itemize}[leftmargin=*]
\item \textbf{Privacy performance penalty.} Federated methods in general accept a performance penalty due to privacy considerations. However, since in our method clients have unfettered access to each other's models, they might adopt more stringent privacy-preserving techniques than they would otherwise. 

\item \textbf{Scaling costs.} Fetch, storing, and making inferences with multiple models from the federation increases the bandwidth, storage, and compute time costs of our technique as models grow in size and number. Thus although clients could smartly select a subset of models to fetch or apply compression techniques, we suggest applying our method with smaller to moderately-sized models when such cost considerations are significant.
\end{itemize}

The rest of the paper is structured as follows. We review related works in Section \ref{sec:related} and explain the details of our proposal in more detail in Section \ref{sec:method}. In Section \ref{sec:experiments} we evaluate our method on synthetic partitions of well-known datasets to simulate different non-IID scenarios. Finally, we conclude and point out future research avenues in Section \ref{sec:conclusion}.

\section{Related Work}
\label{sec:related}

Several personalization strategies for Federated Learning have been proposed. We give a brief overview of the main approaches and then focus on those most relevant to our proposed method. For a more detailed review, we direct the readers to \cite{ludwig_personalization_2022}.

The majority of previous approaches require neural networks or other gradient-based models to be employed at a local and global level. One straightforward technique is fine-tuning or transfer learning, where the global model undergoes additional training epochs using local client data \cite{ludwig_personalization_2022}. Other approaches \cite{shen_federated_2020,li_fedmd:_2019} involve model distillation, which aims to transfer knowledge from the global model to the local models by training the local "student" model to mimic the behavior of the global "teacher" model. 

Additionally, methods inspired by meta-learning, which train models to adapt quickly to new tasks, have been suggested. By treating clients as individual tasks, meta-learning techniques can be utilized to train a global model that readily adapts to the unique characteristics of each client's local data \cite{chen_federated_2019,khodak_adaptive_2019}.

Other personalization techniques can be applied irrespective of the types of local and global models. Client clustering techniques \cite{mansour_three_2020,ghosh_robust_2019,ghosh_efficient_2021} assume that there exists a fixed number of distinct data distributions and attempt to cluster clients with similar distributions to train one “global” model per cluster. 

Data augmentation strategies \cite{mansour_three_2020,mariani_bagan:_2018} generate synthetic data to augment and balance the datasets across clients with the hope of making the client’s datasets “more independent and identically distributed”. Methods where clients send a small fraction of their data to a global data pool have also been proposed \cite{zhao_federated_2018} even though they raise privacy concerns.

Stacked generalization has been applied broadly in Federated Learning. For example, Guo et al. \cite{yao_privacy-preserving_2019} introduced \textit{Privacy Preserving Stacking}, where each client trains a model with privacy guarantees, such as \textit{Privacy-Preserving Logistic Regression (PLR)} \cite{private_log_reg}. Afterward, clients perform predictions on a separate set of their local data, which they send, along with the ground-truth labels, to a central aggregator. The aggregator receives the (prediction, ground-truth) pairs from all the clients and trains a meta-model using these pairs, once again with privacy guarantees. 

Similarly, \textit{FedStack} \cite{safri_federated_2022} proposes training a hierarchical meta-model. A first meta-model is trained on (prediction, ground-truth) pairs received from clients, a second meta-model on pairs outputted by the first-level model, and so on. The complete meta-model is then sent to the clients for inference.

Both \textit{Privacy Preserving Stacking} and \textit{FedStack} approaches result in a global model, and thus do not perform personalization. In contrast, \textit{BaggingWrapper} \cite{wu_wrapperfl:_2022} proposes that clients send each other local models and use them to train a stacked linear model on private data.

Initially introduced as a model-agnostic plug-in technique for clients to join a federation at minimal cost, \textit{BaggingWrapper} was not intended as a personalization technique. Consequently, it was evaluated on a global test set. We propose recasting the core idea as a personalization method and thus will have each client evaluate their personalized models on private test data. 

In the original \textit{BaggingWrapper} design, it was assumed that clients had limited data, resulting in the training of both the private and stacked meta-models on the same dataset. However, this approach risks overfitting stacked models to predictions generated by their respective private models. Therefore, when clients possess sufficient data, we propose training the personalized model on a separate held-out set of private data and evaluate both approaches in our experiments.

In addition, we contribute a simple adaptation to handle vertical and hybrid data partitioning scenarios and leverage the stacked model’s feature importance data as a straightforward way to perform Contribution Evaluation (CE).

\section{Method}
\label{sec:method}

Our method is fairly straightforward and is summarized in Figure \ref{fig:method_fig}. Clients train a privacy-preserving model (PP) on private data and make it available to the federation. Clients can then fetch models from the federation and use them in conjunction with their private non-PP model as base models for training a personalized stacked meta-model. The stacked meta-model is trained on the predictions of the base models on private data.

Training the stacked model on the same data as the private model risks overfitting on its predictions and ignoring the fetched model's predictions. On the other hand, training the personalized model on a held-out set will decrease the data available to train the private model. In general, we recommend training on a held-out set unless the performance of the private models is significantly affected. Of course, clients can find the approach that works best for them by using cross-validation as we describe at the end of the section. 

In horizontal data partitioned federations where every client's data contains the features it is straightforward to obtain predictions from a fetched model. However, in vertical or hybrid federations clients will have different feature sets and will have to match a fetched model's feature set to obtain predictions from it. 

For example, client A may have data with features $\{1,2,3\}$ and wish to make predictions using client B's public model trained with features $\{3, 4, 5\}$. We propose that A simply drop features the fetched model does not contain (in this case $1$ and $2$) and create synthetic features for those the fetched model requires but A does not have (in this case $4$ and $5$).

Several approaches for generating the synthetic features are possible and depend on the metadata each client makes public along with its model. For example, clients could make public which encoding they used for missing data during training so others can use it at prediction time. For simplicity, in our experiments, we assume clients share a set of default feature values that others can use as constant synthetic features. 

To avoid free-loaders and implement fairness mechanisms in the federation, we propose that clients use feature-importance methods on their personalized models. Since a stacked model's features are predictions from fetched models, a client can interpret feature importance as a proxy for how much each model contributed to its personalized model. 

Clients could then choose to use contribution scores in several ways. For example, they could restrict their public models to those with a score above a threshold or give access to versions of their model with different privacy budgets.

In our experiments, we assume that clients make their normalized contribution scores public to the federation where they are aggregated into a weighted directed "contribution graph." Then, any node centrality metric can be used to summarize a client's overall contribution to the federation. 

Finally, we point out that clients are free to perform cross-validation with their private data to evaluate the personalized model and decide whether to train the stacked model on a held-out set or not, how much data to hold out and perform model selection and hyperparameter tuning for both their private and stacked models.

\section{Experiments}
\label{sec:experiments}

We evaluated our method using synthetic partitions of well-known classification datasets (\textit{Census} \cite{misc_census_income_20}, \textit{Cover Type} \cite{misc_covertype_31}, \textit{Vehicle Loan Default} \cite{noauthor_l&t_nodate_2}) to simulate various non-IID scenarios.

Clients trained their public model on their entire private data and then exchanged them. Subsequently, each client randomly split its private data into \verb|(Train, Meta Train, Test)| sets. In our "held-out" approach clients train their private model on \verb|Train| and the meta-model on \verb|Meta Train|, as shown in Figure \ref{fig:method_fig}. In contrast, in the \textit{BaggingWrapper}'s "pooled" approach the private and meta-model are trained on \verb|Train| $\cup$ \verb|Meta train|. Both meta-models are evaluated on the same \verb|Test| set.

The \verb|(Train, Meta Train, Test)| splits were stratified based on the target class and consisted of proportions $0.6$, $0.2$, and $0.2$ of a client's data, respectively. To minimize the impact of random fluctuations in the splitting process, clients repeated the split, training, and evaluation 5 times on the same private data. Furthermore, the random data assignments described in Sections \ref{sec:quantity_skew}, \ref{sec:label_skew}, and \ref{sec:vertical} were repeated on 10 different random seeds per parameter configuration.

For simplicity, clients used Random Forests as their public, private, and personalized meta-models, with default hyperparameters \footnote{Model implementations and default parameters from scikit-learn v1.4.0 \cite{scikit-learn}}. Finally, clients reported the normalized Mean Decrease in Impurity (MDI) of their personalized model as contribution scores to the federation. An individual client's overall contribution (or "importance") to the federation was determined by summing the contribution scores given to them by others and normalizing.

In preprocessing the datasets, we employed one-hot encoding for categorical features, removed examples with missing values, and considered only the two most prominent classes to perform binary classification. For more detailed information regarding data processing, the experimental procedure, and results, refer to the paper's source code \footnote{https://github.com/emiliocantuc/personalized-fl-via-stacking}.

\begin{figure}
  \centering
  \makebox[\textwidth]{\includegraphics[width=\textwidth]{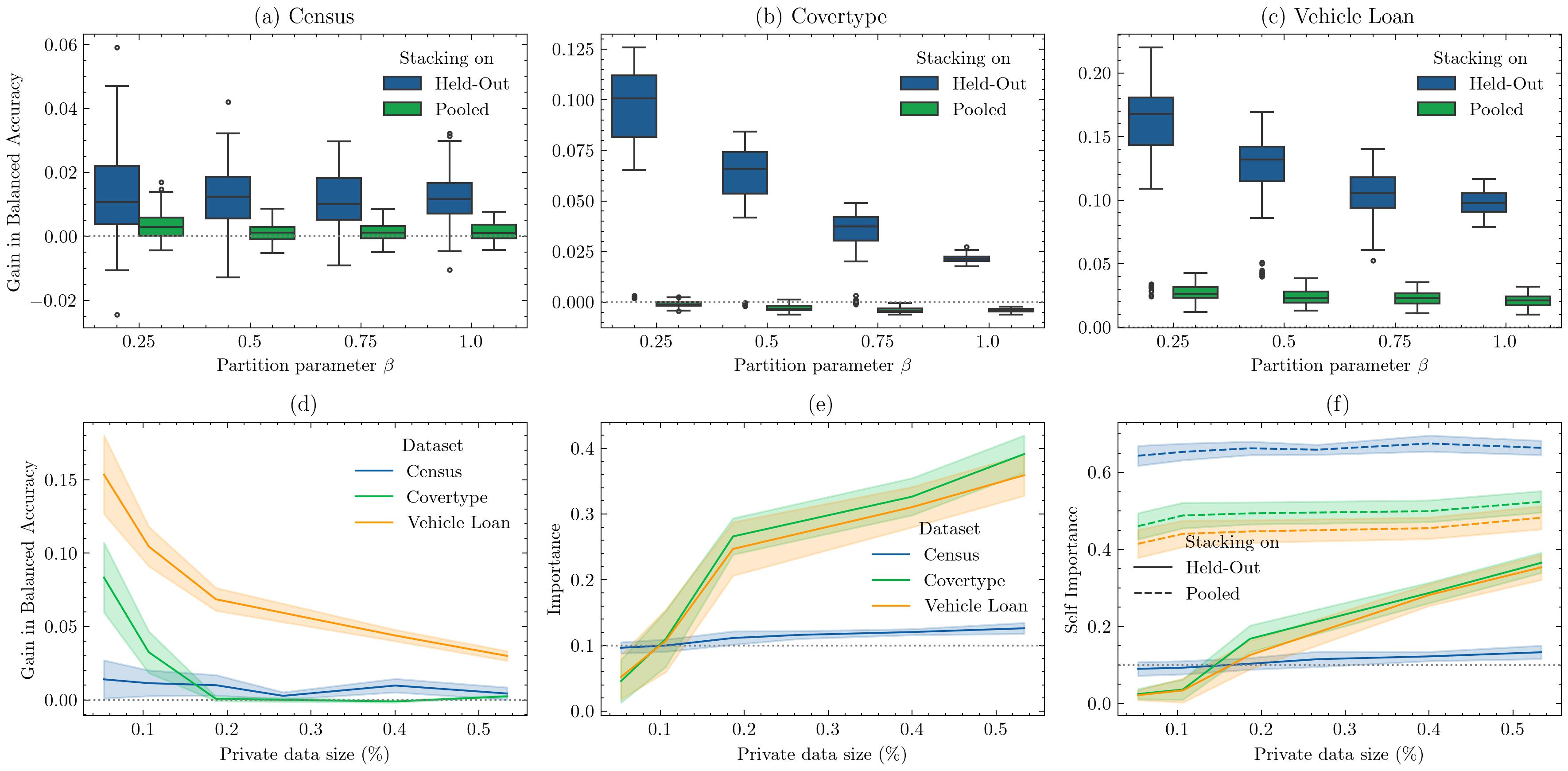}}
  \caption{Quantity skew results: Plots (a)-(c) show the balanced accuracy gain of personalized meta-models over private ones, trained on both held-out and pooled data. Plots (d)-(f) depict performance gain, importance, and self-importance relative to private data size. Note that (d) and (e) focus on meta-models stacked on held-out data.}
  \label{fig:power}
\end{figure}

\subsection{Quantity skew}
\label{sec:quantity_skew}

We initially simulate synthetic partitions where only the quantity of data varies among 10 clients according to a power distribution with a parameter $\beta \in (0,1]$. When $\beta = 1$, the data is evenly distributed among clients, whereas as $\beta$ approaches 0, fewer clients possess a larger share of the data.

Figure \ref{fig:power} (a)-(c) display the improvement in predictive performance of the personalized meta-models over the private models. Notably, across datasets, training the personalized model on the held-out \verb|Meta Train| set results in stronger performance compared to training both the private and meta-model on the pooled \verb|Train| $\cup$  \verb|Meta Train| set.

Furthermore, as we increase $\beta$ and distribute the data more evenly among clients, the gain from personalization tends to decrease. However, the \textit{Census} dataset shows consistent performance regardless of the distribution, likely due to its lower data requirement for achieving good results. 

In (c) and (e), we observe that as clients possess a larger share of the data, their performance gain from personalization diminishes, but their perceived contribution to the federation ("importance") increases. 

Finally, (f) presents the importance that clients' stacked models assign to their private ones, referred to as their "self-importance." We note that stacking on the pooled dataset leads to meta-models assigning a higher weight to private models overall, likely overfitting on them. When stacking on a held-out set the meta-models learn to assign greater weight to the private models as they are trained on more data. 

\subsection{Label skew}
\label{sec:label_skew}

To simulate label skew, representing a change in $P(Y|X)$ across 10 clients, we adopt the approach outlined in \cite{hsu2019measuring}. We independently assign the same number of examples to each client such that its target labels follow a categorical distribution over $K$ classes with probabilities $\mathbf{q} = (q_1, ..., q_k)$ (with $q_i \geq 0$ and $\Vert \mathbf{q} \Vert_1 = 1$). The class probabilities are obtained by sampling $\mathbf{q} \sim Dirichlet(\alpha\mathbf{p})$, where $\mathbf{p}$ is the prior class distribution and $\alpha > 0$ controls the similarity among clients. When $\alpha \to 0$ clients are likely to have distinct and highly imbalanced labels and as $\alpha \to \infty$ the label distribution will converge to the prior $\mathbf{p}$ for all clients.

\begin{figure}
  \centering
  \makebox[\textwidth]{\includegraphics[width=\textwidth]{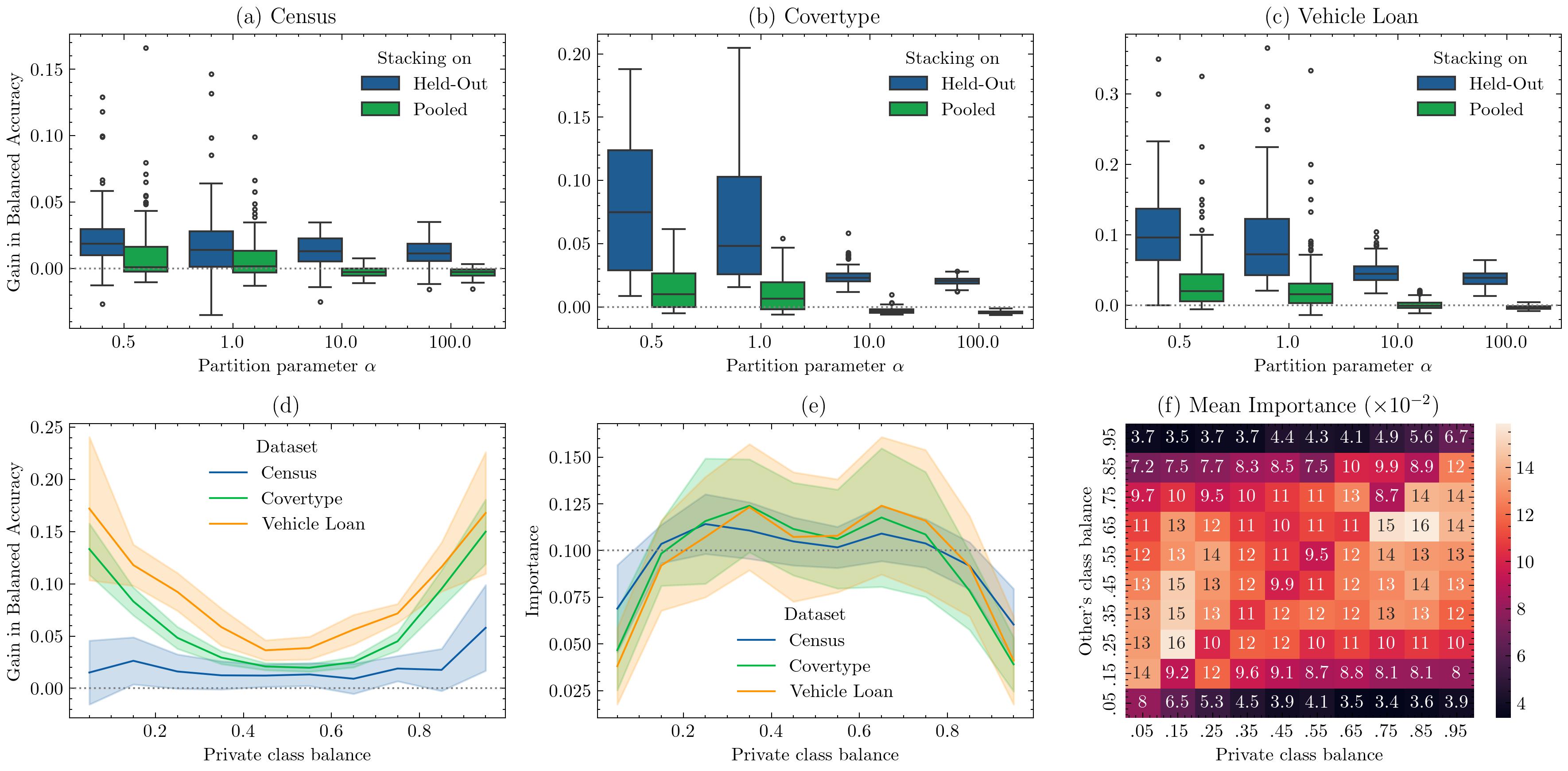}}
  \caption{Label skew results. Plots (a)-(c) show balanced accuracy gain. Plots (d) and (e) depict performance gain and importance relative to clients' private label balance, respectively. Plot (f) displays assigned importance as a function of both a client's and another's private label balance. Note that (d)-(f) focus on meta-models stacked on held-out data.}
  \label{fig:dirichlet}
\end{figure}

Similar to the quantity skew scenarios, Figure \ref{fig:dirichlet} (a)-(c) illustrates that stacking on a held-out set achieves a higher performance gain across datasets and different values of $\alpha$. Furthermore, we note that as $\alpha$ increases and the data among clients becomes more class-balanced, the gains from personalization diminish.

In (d) we observe that, in general, clients tend to experience performance gains through personalization as their private data becomes more class-imbalanced. Although clients with class-balanced data may not gain as much as others, our contribution metric appears to effectively capture their increased importance to the federation, as depicted in Figure (e).

Lastly, (f) displays the average importance a client learns to assign to others as a function of both label balances. We observe that while generally class-balanced clients tend to get assigned higher importance, clients have a bias for others with similar but less extreme imbalances. For instance, clients with positive label proportions around 0.15 tend to prioritize models trained with proportions between 0.25 and 0.35, rather than more balanced ones. It is also worth noting that personalized models tend to assign less weight to models trained with similar class imbalances as the private model, as exhibited by the relatively lower entries in the diagonal.

\subsection{Vertical partitioning}
\label{sec:vertical}

To evaluate the performance of our method on vertically partitioned data we assign to each of 10 clients all the examples but only a proportion $p$ of predictors selected uniformly at random with replacement. 

As outlined in Section \ref{sec:method}, clients publicly provide a set of default values for each feature so that others can impute missing values at prediction time. For categorical variables, we have clients share their most frequent value, while for continuous variables, they share their median with added Gaussian noise. Specifically, for a continuous feature $k$ and client $c$, the default value shared is $med_{c,k} + N(0, s_{c,k} \times \varepsilon_{c,k})$, where $s_{c,k}$ is the feature $k$'s sample standard deviation for client $c$. For our simulations, we consider the case where $\varepsilon_{c,k} = \varepsilon$ for simplicity.

\begin{figure}
  \centering
  \makebox[\textwidth]{\includegraphics[width=\textwidth]{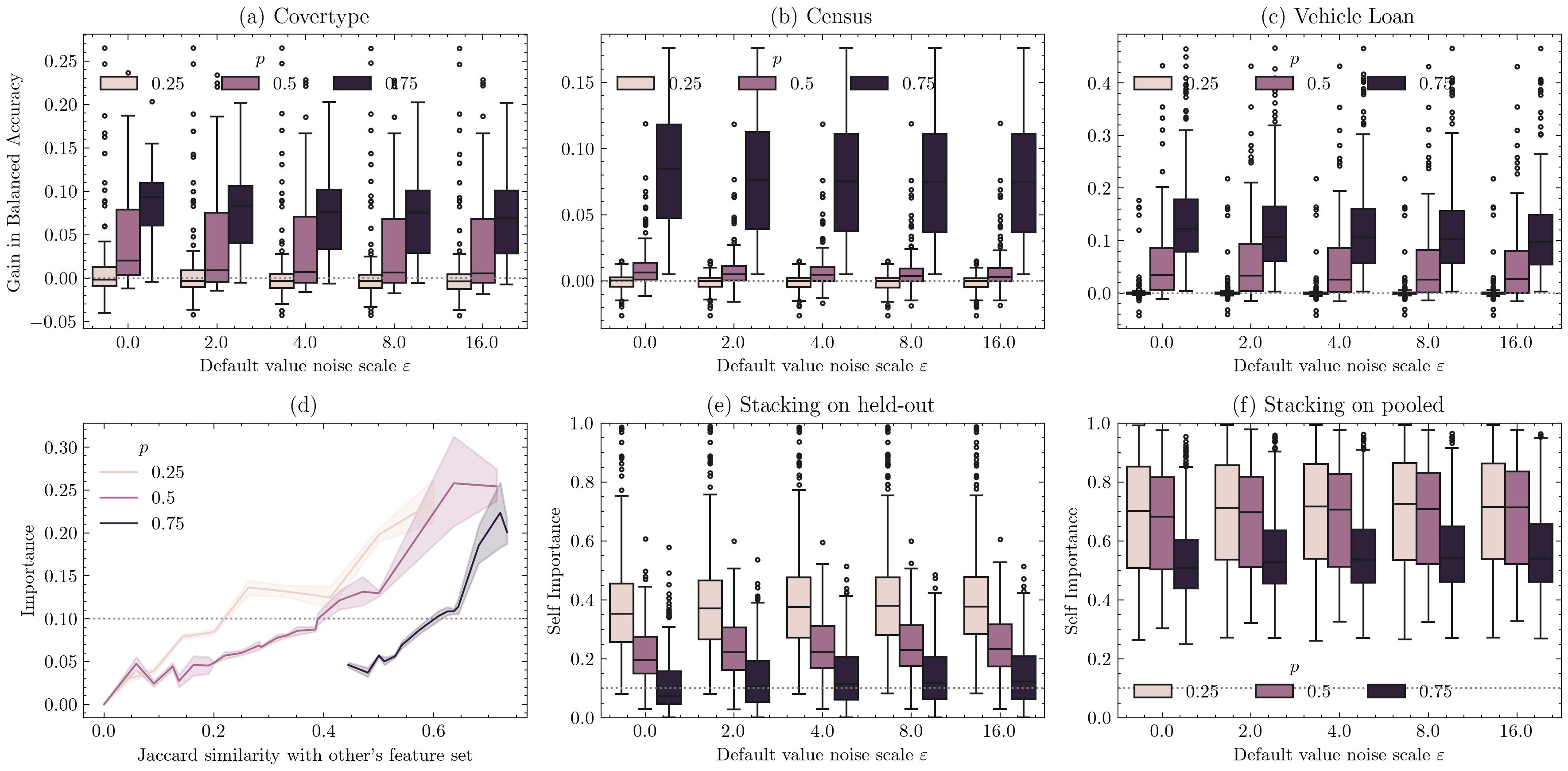}}
  \caption{Vertical partitioning results. Plots (a)-(c) depict balanced accuracy gain based on the proportion $p$ of total features randomly assigned to clients and the noise level $\varepsilon$ added to default values. Plot (d) illustrates assigned importance as a function of Jaccard similarity between feature sets. Plots (e) and (f) show self-importance relative to varying $p$ and $\varepsilon$, for both meta-models stacked on held-out and pooled data. Note that plots (a)-(d) focus on meta-models stacked on held-out data.}
  \label{fig:vertical}
\end{figure}

Figure \ref{fig:vertical} (a) - (c) display the performance gain by personalization as a function of $p$ and $\varepsilon$.  We present results only for personalized models stacked on held-out data, as those stacked on pooled sets had negligible gains (see Figure \ref{fig:vertical_appendix} in Appendix A).

As $p$ increases and clients share more predictors, we observe that they tend to assign less weight to their private models (see (f) and (e)) and benefit more from personalization. This trend aligns with our intuition, as clients with fewer common features will have to impute more noisy default values. In addition, we observe that as $\varepsilon$ increases and the default values become noisier, personalized models tend to rely more on their private models which results in diminished performance gains.

Finally, in Figure (d), we present the assigned importance that a client gives to another's model as a function of the Jaccard Similarity of their feature sets. We observe a strong relationship, indicating that clients learn to assign increasing weight to models with a greater degree of overlap in their feature sets.

\subsection{Natural partitioning}

\begin{figure}
  \centering
  \makebox[\textwidth]{\includegraphics[width=\textwidth]{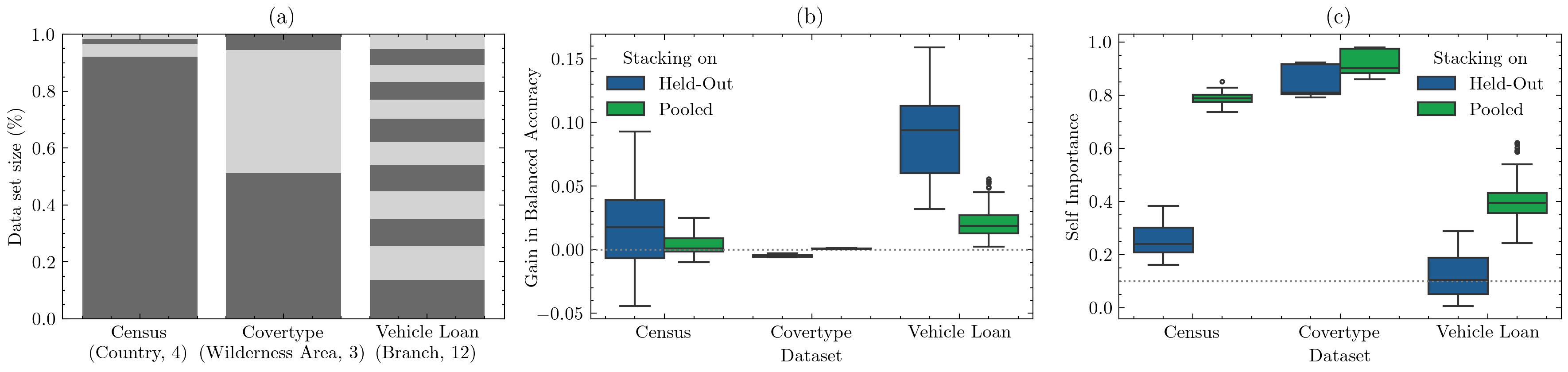}}
  \caption{Natural partitioning results. Plot (a) displays the column used for dataset partitioning, along with the resulting number of clients and their proportion of total data. Plots (b) and (c) show the performance gain and self-importance of meta-models stacked on both held-out and pooled data across datasets.}
  \label{fig:natural}
\end{figure}

Finally, we aim to mimic realistic data partitions commonly seen in real-world scenarios by partitioning a categorical column within our datasets. Figure \ref{fig:natural} (a) gives an overview of the specific column each dataset was partitioned on, alongside the resulting number of clients and the distribution of data quantity. For instance, in the \textit{Cover Type} dataset, we treat each of the three Wilderness Areas as a separate client and can observe that two clients hold the majority of the data.

Upon analyzing Figures (b) and (c), we notice that in the \textit{Census} and \textit{Vehicle Loan} datasets, most clients benefited from personalization, particularly when stacking on held-out data. Similar to previous findings, stacking on the pooled set tended to favor private models, leading to performance degradation. However, in the \textit{Cover Type} dataset, personalization did not result in significant gains overall, and stacking on a held-out set even led to reduced performance. This underscores the significant impact of data nature and distribution across clients on potential personalization gains.

\section{Discussion and Future Work}
\label{sec:conclusion}

This paper introduced a straightforward yet effective Personalized Federated Learning method based on stacked generalization. The proposed approach is simple, flexible, and applicable to horizontal, hybrid, and vertically partitioned federations. Moreover, our method provides a natural means to evaluate each client's contribution to the federation.

Extensive evaluation under simulated data heterogeneity demonstrated the method's effectiveness and intuitive behavior. Notably, we observed that, in general, training the personalized meta-model on the same data as a client's private model often led to reduced performance gains due to overfitting.

While our experiments provided valuable insights, there are several interesting potential research directions worth exploring. Firstly, investigating the impact of various privacy-preserving techniques would be valuable. Additionally, in vertically partitioned scenarios, exploring alternative methods of constructing default values and examining the effects of added noise would be beneficial.

Moreover, developing novel mechanisms to utilize contribution scores for fair collaboration and decision-making within federated settings deserves attention. Lastly, advancing the theoretical understanding of generalized stacking on non-IID data will be essential in improving the robustness and scalability of personalized Federated Learning methods.

\bibliographystyle{unsrt}  
\bibliography{references}  

\newpage
\section*{Appendix A}
\label{sec:appendix}

\begin{figure}[hbt!]
  \centering
  \makebox[\textwidth]{\includegraphics[width=\textwidth]{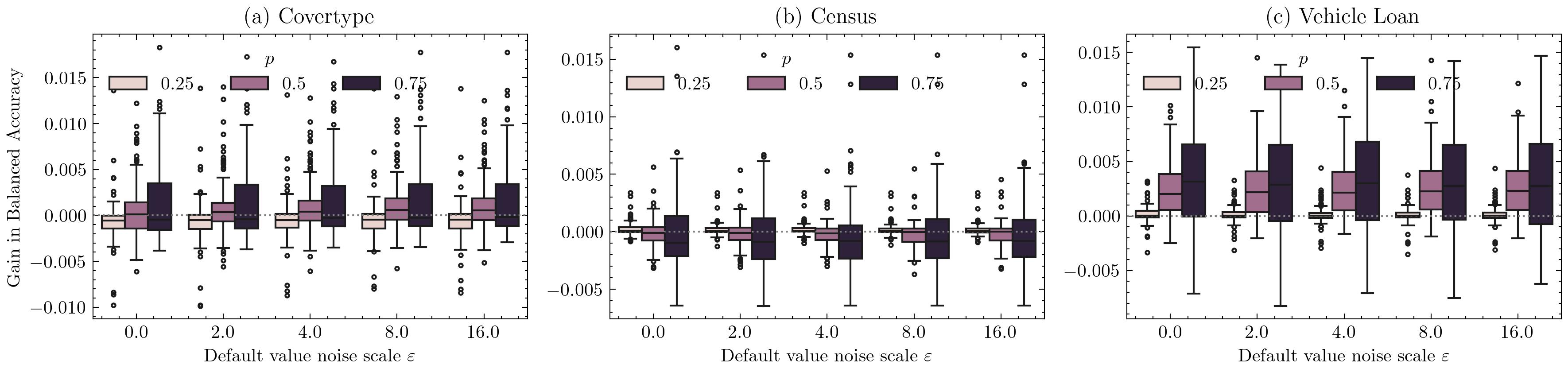}}
  \caption{Vertical partitioning gain in performance when personalized models are trained on the same (pooled) data as private models.}
  \label{fig:vertical_appendix}
\end{figure}

As mentioned in Section \ref{sec:vertical}, Figure \ref{fig:vertical} showcases the performance gains for meta-models stacked on held-out data. In contrast, Figure \ref{fig:vertical_appendix} demonstrates the comparatively negligible gains for models stacked on the same data as private models.

\end{document}